\documentclass[10pt,twocolumn,letterpaper]{article}

\usepackage[pagenumbers]{iccv} % To force page numbers, e.g. for an arXiv version
\usepackage{booktabs}
\usepackage{multirow}
\usepackage{url}
\usepackage{multicol}
\usepackage[normalem]{ulem}
\useunder{\uline}{\ul}{}
\usepackage{algorithm}
\usepackage{algpseudocode}
\usepackage[accsupp]{axessibility}
\usepackage{flushend}

\definecolor{iccvblue}{rgb}{0.21,0.49,0.74}
\usepackage[pagebackref,breaklinks,colorlinks,allcolors=iccvblue]{hyperref}

\def\paperID{*****} % *** Enter the Paper ID here
\def\confName{ICCV}
\def\confYear{2025}

\title{\LARGE \bf
FUSELOC: Fusing Global and Local Descriptors for\\ Fast and Robust 2D-3D Matching in Visual Localization
}

\author{Son Tung Nguyen \qquad Alejandro Fontan \qquad Michael Milford \qquad Tobias Fischer\\[0.25cm]
Queensland University of Technology\\
Brisbane, Australia\\
{\tt\small sontung.nguyen@hdr.qut.edu.au} \\
}

\DeclareMathOperator*{\argmin}{arg\,min}

\newcommand{\desc}{\mathbf{d}}
\newcommand{\point}{\mathbf{p}}

\newcommand{\localdesc}{\mathbf{d}^{\text{local}}}
\newcommand{\globaldesc}{\mathbf{d}^{\text{global}}}

\newcommand{\extrinsic}{H}

\begin{document}
\maketitle
\begin{abstract}
Hierarchical visual localization methods achieve state-of-the-art accuracy but require substantial memory as they need to store all database images. Direct 2D-3D matching requires significantly less memory but suffers from lower accuracy due to the larger and more ambiguous search space. We address this ambiguity by fusing local and global descriptors using a weighted average operator. This operator rearranges the local descriptor space so that geographically nearby local descriptors are closer in the feature space according to the global descriptors. This decreases the number of irrelevant competing descriptors, especially if they are geographically distant, thus increasing correct matching likelihood. We consistently improve the accuracy over local-only systems and we achieve performance close to hierarchical methods while using 43\% less memory and running 1.6 times faster. Extensive experiments on four challenging datasets -- Cambridge Landmarks, Aachen Day/Night, RobotCar Seasons, and Extended CMU Seasons -- demonstrate that, for the first time, direct matching algorithms can benefit from global descriptors without compromising computational efficiency. Our code is available at \href{https://github.com/sontung/descriptor-disambiguation}{https://github.com/sontung/descriptor-disambiguation}.
\end{abstract}

\section{Introduction}
\label{sec:intro}

% introduce problems
Visual localization is the process of determining the pose (position and orientation) of a camera or a robot within its environment by analyzing visual information obtained from RGB images. This typically involves comparing observed features in camera images against a pre-existing reference point cloud (referred to as the \textit{map}) to estimate the camera pose. Visual localization enables effective navigation using only visual cues, rendering it particularly valuable in environments where GPS signals may be unreliable or unavailable, such as indoor spaces or densely built urban areas.
% Visual localization is a key component of various applications such as augmented reality~\cite{arth2011real, castle2008video}, autonomous vehicles~\cite{heng2019project, lim2012real}, and indoor navigation systems~\cite{moller2012mobile}, enabling effective navigation using only visual cues. This makes visual localization particularly useful in environments where GPS signals may be unreliable or unavailable, such as indoor spaces or densely built urban areas.

Several classes of solutions address the visual localization problem, each with distinct strengths and weaknesses. Among these, direct 2D-3D matching~\cite{DBLP:conf/eccv/LiSH10, active_search, sattler2011fast} and hierarchical solutions~\cite{hloc, peng2021megloc} are notable for their accuracy in large-scale outdoor maps, from small buildings to entire cities. Hierarchical solutions achieve robust performance by using image retrieval systems~\cite{hausler2021patch, DBLP:conf/cvpr/ArandjelovicGTP16, ali2023mixvpr, berton2023eigenplaces, tolias2013aggregate} to identify similar database images for feature matching. This process serves as a coarse pose estimation, guiding the search to relevant regions of the map and reducing search ambiguity. 

Hierarchical solutions benefit greatly from advances in image retrieval systems~\cite{ali2023mixvpr, berton2023eigenplaces, Izquierdo_CVPR_2024_SALAD} and represent state-of-the-art solutions for visual localization. However, their accuracy comes with substantial memory requirements, as all database images and global descriptors must be stored. In contrast, direct 2D-3D matching systems require approximately half the memory for city-scale maps (see Table~\ref{tab:results_combined}).

The main drawback of direct matching algorithms is caused by perceptual aliasing in large-scale maps, which creates search space ambiguity and results in numerous false matches between query pixels and the point cloud. To address this, we draw inspiration from hierarchical methods and integrate robust image retrieval techniques to enhance local descriptors during search operations. Specifically, we fuse global descriptors with local descriptors through feature averaging. Compared to standard 2D-3D search algorithms, the only additional memory overhead is that of the retrieval network's weights, as the feature descriptor size remains unchanged. Despite its simplicity, our fused descriptors significantly reduce search ambiguity, leading to notable accuracy improvements in extensive experiments when integrated into a nearest-neighbor lookup system~\cite{sattler2011fast}.

\begin{figure*}[t]
    \centering
    \includegraphics[width=0.97\linewidth]{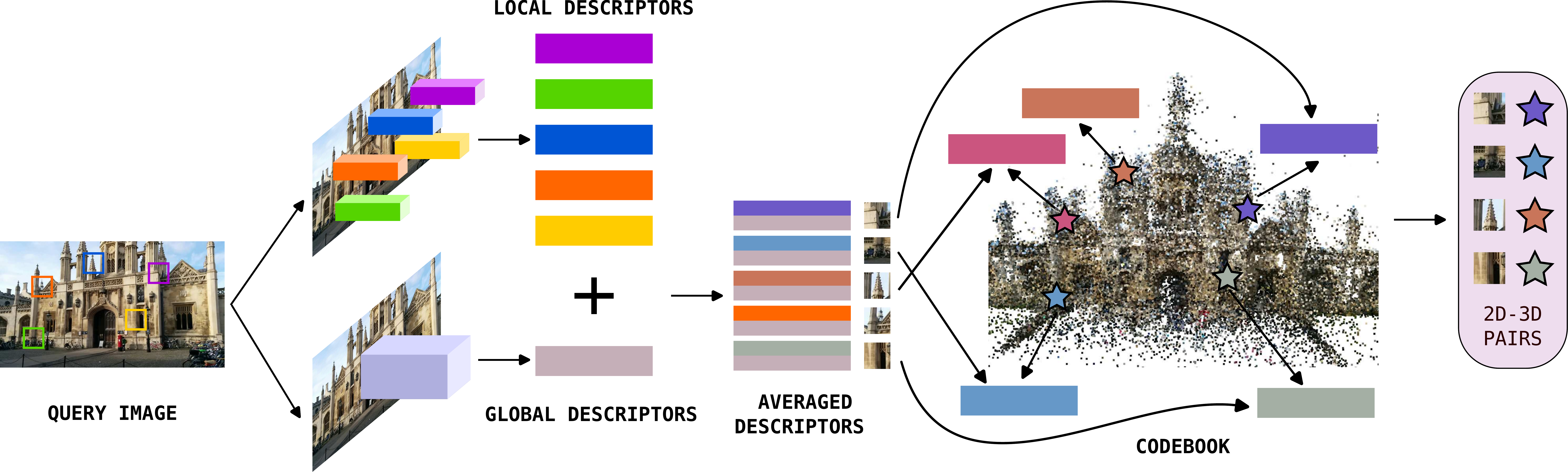}
    \caption{\textbf{System overview.} 
    Inspired by hierarchical visual localization methods~\cite{hloc, peng2021megloc}, we integrate global descriptors into a direct 2D-3D matching baseline to reduce search space ambiguity. First, global and local descriptors for the query image are obtained. These descriptors are fused using a weighted average operator (+). The fused descriptors are then used to perform nearest-neighbor searches against the database codebook to establish 2D-3D pairs. This design minimally increases the computational overhead due to the retrieval system while significantly enhancing accuracy compared to conventional 2D-3D search systems.
    }
    \label{fig:system_overview}
\end{figure*}

\noindent We summarize our contributions as follows:
\begin{enumerate}
    \item We integrate image retrieval techniques into direct 2D-3D matching systems by employing a weighted average operator to combine global and local descriptors in nearest-neighbor lookup (Figure~\ref{fig:system_overview}).
    \item We conduct extensive experiments using four large-scale outdoor datasets~\cite{kendall2015posenet,zhang2021reference,Maddern2017IJRR,sattler2018benchmarking} to demonstrate the significant positive impact of our design on accuracy without adversely affecting computational complexity.
    %\item We perform comprehensive ablation studies to provide an in-depth understanding of the optimal settings for our system.
    \item We perform comprehensive ablation studies to analyze the sensitivity of our system settings and demonstrate that a wide range of weightings for local and global descriptors consistently outperforms local-only approaches.
\end{enumerate}

% reason over the improvement of hloc over the years, because of retrieval methods
% why should we care about focus on 2d-3d matching

\section{Related works}
\label{sec:rw}

We begin by reviewing direct matching solutions (Section~\ref{sec:direct}), followed by hierarchical solutions (\mbox{Section~\ref{sec:hierarchical}}) and their crucial components, namely global (\mbox{Section~\ref{sec:global}}) and local (Section~\ref{sec:local}) descriptors. We then review learning-based solutions (Section~\ref{sec:learning}). Finally, we review other methods that combine global and local descriptors in image retrieval and visual localization (Section~\ref{sec:combine}).

\subsection{Direct 2D-3D matching for visual localization} \label{sec:direct}
Early solutions to the visual localization problem~\cite{DBLP:conf/eccv/LiSH10, active_search, sattler2011fast} focused on directly matching 2D features with a point cloud to establish 2D-3D correspondences. These methods typically construct a descriptor codebook for each point in the point cloud by averaging the descriptors of all database pixels in which the points are visible. 
Although 2D-3D matching algorithms are more memory-efficient, they fall short of hierarchical approaches on challenging datasets~\cite{sattler2012image, sattler2018benchmarking, cmu_dataset, Maddern2017IJRR} due to the large and ambiguous nature of the search space. To address these limitations, we propose integrating global descriptors into a direct matching framework, achieving superior performance compared to local descriptor-only techniques while retaining the simplicity and efficiency inherent to direct matching methods.

\subsection{Hierarchical visual localization} \label{sec:hierarchical}
Visual localization problems in outdoor scenes often involve a vast search space~\cite{Maddern2017IJRR, cmu_dataset}. To address this challenge, several approaches~\cite{peng2021megloc, hloc} have proposed leveraging image retrieval techniques to streamline and refine the search process, enhancing its efficiency and accuracy. These approaches typically begin by retrieving database images similar to the query image, and then establish 2D-2D feature correspondences between these retrieved images and the query image to establish 2D-3D correspondences. Despite achieving state-of-the-art accuracy, such methods often demand significant memory resources as they require access to all database images alongside the point cloud coordinates. On the other hand, direct matching algorithms bypass the 2D-2D feature matching step, thus requiring no database images and being a lot more memory efficient. 

Recent work~\cite{wang2024mad, laskar2024differentiable} shows that compressing descriptors enables lightweight yet accurate hierarchical systems. However, we argue that direct matching methods can also benefit from such compression, as the descriptor codebook constitutes the dominant portion of their memory usage. Therefore, to ensure a fair comparison, we present our analysis under the condition that both systems are used in their original, uncompressed form. 

In this work, we propose enhancing the performance of direct matching by incorporating image retrieval, as in hierarchical methods, while preserving its low memory footprint.

\subsection{Image retrieval for visual localization} \label{sec:global}
Image retrieval is the task of finding the most similar images to the input image~\cite{masone2021survey, DBLP:journals/trob/LowryS0LCCM16}. Current systems often reduce this problem to similarity search in a $d$-dimensional space~\cite{hausler2021patch, DBLP:conf/cvpr/ArandjelovicGTP16, ali2023mixvpr, berton2023eigenplaces, DBLP:conf/mir/MohedanoMOSMN16, tolias2013aggregate}. Therefore, a similarity function must be established to compare any given pair of images using their $d$-dimensional global descriptors~\cite{schubert2023visual}. This can be done by aggregating either local descriptors~\cite{DBLP:conf/mir/MohedanoMOSMN16, tolias2013aggregate}, multiple convolutional neural network layers~\cite{hausler2021patch, DBLP:conf/cvpr/ArandjelovicGTP16, ali2023mixvpr, berton2023eigenplaces}, or DINO features~\cite{Izquierdo_CVPR_2024_SALAD} into a single global descriptor vector. Image retrieval helps to reduce the correspondence search space, which is crucial for hierarchical visual localization~\cite{hloc, peng2021megloc}. However, retrieval systems require access to the database global descriptors, contributing to hierarchical systems' high memory usage. In this paper, we propose integrating image retrieval methods to disambiguate the 2D-3D correspondence search space without storing global descriptors.

\subsection{Local descriptor for visual localization} \label{sec:local}
Classical local feature methods~\cite{DBLP:journals/ijcv/Lowe04, DBLP:conf/eccv/BayTG06} detect invariant pixels that can be tracked across viewpoints. These methods are fast in practice and perform very well in real-world scenarios, thus they are commonly deployed in structured localization systems~\cite{DBLP:conf/eccv/LiSH10, active_search, DBLP:conf/cvpr/IrscharaZFB09, sattler2016efficient}. 
Recent works~\cite{revaud2019r2d2, dusmanu2019d2, detone2018superpoint} proposed to use deep networks to learn both feature detection and description. SuperPoint~\cite{detone2018superpoint} presented a self-supervised framework tailored for training interest point detectors and descriptors, thus eliminating the need to define interest points manually.
D2~\cite{dusmanu2019d2} uses a single convolutional neural network that serves for both dense feature description and feature detection. By deferring detection to a later stage, the resulting keypoints exhibit greater stability than those from traditional methods that rely on early detection of low-level structures.
R2D2~\cite{revaud2019r2d2} simultaneously learns keypoint detection and description, along with a predictor for local descriptor discriminativeness. This approach aims to mitigate ambiguous areas, resulting in more reliable keypoint detection and description.
Our paper explores highly-performing local feature detectors and demonstrates that fusing them with global descriptors enhances their performance in 2D-3D correspondence search.

\subsection{Learning-based visual localization} \label{sec:learning}

Deep learning has enabled novel solutions for visual localization. Absolute pose regression models directly output camera poses for a query image~\cite{kendall2015posenet, DBLP:conf/iccv/ShavitFK21}. They are fast, but their lack of efficient optimization leads to reduced performance. On the other hand, scene coordinate regression models that output scene coordinates for query pixels~\cite{DBLP:conf/cvpr/ShottonGZICF13, DBLP:journals/pami/BrachmannR22, brachmann2023accelerated, brachmann2018learning, GLACE2024CVPR, focustune} can be optimized effectively using the re-projection error, perform well in practice, and achieve significantly higher accuracy compared to absolute pose regression models.
DUSt3R~\cite{duster} produces dense 2D-3D mappings for unconstrained image collections and demonstrates impressive performance in 3D reconstruction, as well as in absolute and relative pose estimation~\cite{chen2024map, arnold2022map}. 
Overall, learning-based methods have achieved remarkable improvement; however, their performance is still not on par with hierarchical or structured methods on large-scale maps.

\subsection{Combining global and local features} \label{sec:combine}
Numerous studies~\cite{cao2020unifying, yang2021dolg, phan2022patch, wu2023asymmetric, loquercio2017efficient, kim2015predicting} have highlighted the benefits of combining global and local features to enhance image retrieval systems. Global features provide robustness to changes in viewpoint and lighting, while local features capture detailed geometry and texture, making their combination beneficial for improved performance. However, existing combination methods tend to be computationally demanding~\cite{cao2020unifying, yang2021dolg}, complex~\cite{phan2022patch, wu2023asymmetric}, or require retraining~\cite{phan2022patch, wu2023asymmetric}, limiting their suitability for large-scale visual localization where efficiency is essential. The work most similar to ours, GLACE~\cite{GLACE2024CVPR}, integrates global and local descriptors in ACE~\cite{brachmann2023accelerated} via concatenation to create an accurate scene coordinate regressor. This approach, however, results in a high-dimensional output that increases the database codebook size linearly, making it impractical for structure-based methods. Additionally, its inability to disambiguate similar-looking objects renders it inapplicable for large-scale environments. In contrast, we demonstrate that a weighted descriptor average significantly enhances performance for direct matching methods within a nearest-neighbor lookup system, while incurring minimal computational complexity.

\section{Preliminaries}
\begin{figure*}[t]
    \centering
    \includegraphics[width=1\linewidth]{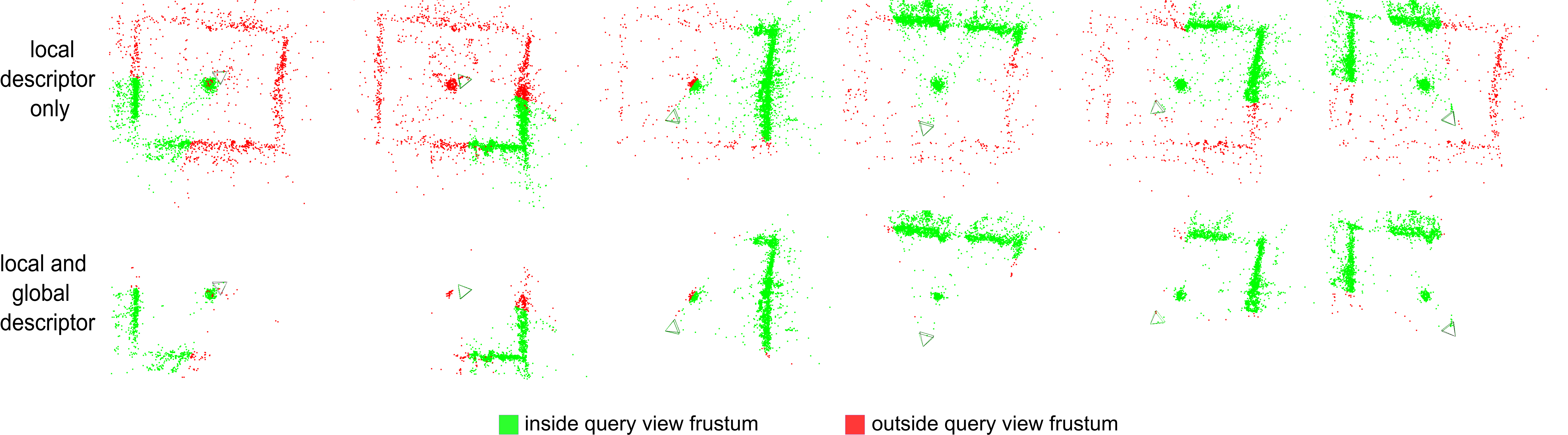}
    \caption{\textbf{Retrieved points using different codebooks.} Retrieved points are visualized in five query images using two different codebooks from the Great Court sequence of the Cambridge Landmarks dataset~\cite{kendall2015posenet}. Points outside the \textit{ground truth} query view frustum (incorrect matches) are shown in \textit{red}, while correct matches are shown in \textit{green}. For each query image, the percentage of correct retrievals is plotted at the top-right corner. The vanilla codebook (top) retrieves many false matches, with points scattered around the map and outside the view frustum. By fusing local and global descriptors, the number of false retrievals is significantly reduced, with most retrieved points appearing within the view frustum (bottom).}
    \label{fig:matches_comparision}
\end{figure*}

%rewritten from the wacv paper
Given a ground truth point cloud map reconstructed using Structure-from-Motion (SfM)~\cite{schoenberger2016sfm} and the associated database images, the visual localization problem is to determine the 6-DoF camera pose $\extrinsic \in \text{SE}(3)$ for a query image with respect to the given point cloud map.

To address this problem, it is essential to establish sufficient correspondences between the pixels of the query image and the point cloud of the provided map. While various methods exist to tackle this problem, as reviewed in Section~\ref{sec:rw}, this paper focuses on the direct 2D-3D matching method due to its favorable memory requirements. Typically, 2D-3D matching employs a codebook that assigns a descriptor to each point within the point cloud map, enabling comparisons against the local descriptors extracted from the query image.

However, a significant limitation of this approach is the ambiguity within the codebook. In environments with repetitive local details, relying solely on local features is inadequate for distinguishing between different areas of the map. Therefore, utilizing a codebook based solely on local descriptors leads to numerous false matches (see Figure~\ref{fig:matches_comparision}, top row), which negatively impacting the final pose estimation.

\section{Methodology}
% summarize baseline system
This section describes the integration of our disambiguated descriptors into a simple nearest-neighbor direct matching system~\cite{sattler2011fast} (Figure~\ref{fig:system_overview}). During training, we generate a descriptor codebook that specifies a descriptor for each point in the reference point cloud (Section \ref{subsec:codebook}). This is accomplished by gathering all local and global descriptors using the database images in which a point appears. The choice of descriptors is discussed in Section~\ref{subsec:descriptors}. At query time, we process the local descriptors and a global descriptor of the query image to match against the codebook, thereby establishing 2D-3D correspondences between the query image's pixels and the 3D points of the point cloud. These correspondences are then fed into a RANSAC-PnP~\cite{PoseLib} loop to compute the camera pose (Section~\ref{subsec:querytime}).

\subsection{Codebook}
\label{subsec:codebook}
Following earlier works~\cite{active_search, sattler2011fast}, we create a codebook for the map. Each entry in the codebook contains a descriptor $\desc_i$ and its associated 3D coordinate $\point_i$. During training, the codebook is constructed by gathering descriptors for all points using the database images. For each point, we assign it the mean descriptor of all its appearance descriptors $\desc_{ij}$ across the database images:
\begin{equation}
\label{eq:codebook}
    \desc_i = \frac{1}{N_i}(\desc_{i1}+\desc_{i2}+\desc_{i3}+ \dots + \desc_{ij} + \dots + \desc_{iN_i}),
\end{equation}
where $\desc_i$ is the descriptor of the $i$-th point in the codebook, $\desc_{ij}$ is the appearance descriptor of $\point_i$ in the $j$-th  database image, and $N_i$ is the number of database images in which $\point_i$ appears. The appearance descriptor $\desc_{ij}$ is computed using:
\begin{equation}
\label{eq:codebook2}
    \desc_{ij} = \lambda \localdesc_{ij} + (1-\lambda)\globaldesc_{j},
\end{equation}
where $\localdesc_{ij}$ is the local descriptor of $\point_i$ in its $j$-th appearance, $\globaldesc_{j}$ is the global descriptor of the $j$-th database image in which $\point_i$ appears, and $\lambda$ controls the contribution of the local and global descriptors to the appearance descriptor.

\subsection{Local and global descriptors}
\label{subsec:descriptors}
We observe that using only local feature descriptors results in highly ambiguous codebooks, leading to numerous false matches (Figure~\ref{fig:matches_comparision}, top row). Global descriptors, however, effectively distinguish different sections of the map, thus reducing codebook ambiguity. Combining global descriptors with local descriptors within our system results in highly effective visual localization. Since global descriptor methods output vectors with varying dimensions, it is necessary to reduce the dimension of the global descriptor to match that of the chosen local descriptor (see Equation~\ref{eq:codebook2}). Section~\ref{subsec:descriptortruncation} discusses various methods for truncating both database and query global descriptors, including random index selection and random Gaussian projections~\cite{bingham2001random}. We use random index selection in all of our experiments.

\begin{table*}[ht]
\centering
\scriptsize
\resizebox{\textwidth}{!}{%
\begin{tabular}{lccccccc}
\toprule
\multicolumn{1}{l}{} &  & \multicolumn{5}{c}{Cambridge Landmarks} \\
\cmidrule(l){3-7} 
\multicolumn{1}{l}{}   & \multirow{-2}{*}{\begin{tabular}[c]{@{}c@{}}Memory\\ requirement $\downarrow$\end{tabular}} & Court & King's & Hospital & Shop & St. Mary's 
& \multirow{-2}{*}{\begin{tabular}[c]{@{}c@{}}Average $\downarrow$ \\ (cm / $^\circ$)\end{tabular}} \\ 

\midrule
  hLoc (SP+SG) \cite{hloc, sarlin2020superglue} &    $\sim$4 GB & {\ul 16/0.1} & {12/0.2} & {15/0.3} & \textbf{4/0.2} & {7/0.2} & {10.8/0.2} \\
 
  AS (SIFT) \cite{active_search}    &  $\sim$200 MB & 24/0.1 & 13/0.2 & { 20/0.4} & \textbf{4.0/0.2} & 8.0/0.3 & 14.0/0.2 \\
  SuperPoint~\cite{detone2018superpoint} & $\sim$49 MB & 28.0/0.1&10.7/0.2&15.1/0.3&{\ul 4.1/0.2}&7.2/0.2&13.0/0.2  \\
  SIFT~\cite{DBLP:journals/ijcv/Lowe04} & $\sim$26 MB & 23.4/0.1&{\ul 10.4/0.2}&{\ul 13.3/0.3}&4.2/0.2&6.8/0.2&11.6/0.2 \\
  R2D2~\cite{revaud2019r2d2} & $\sim$26 MB & 21.6/0.1&\textbf{10.3/0.2}&14.1/0.3&4.3/0.2&{6.6/0.2}&11.4/0.2  \\
  D2~\cite{dusmanu2019d2} & $\sim$244 MB & 24.4/0.1&10.7/0.2&13.8/0.3&4.6/0.2&6.8/0.2&12.1/0.2 \\

  GLACE \cite{GLACE2024CVPR} & 13 MB & {19/0.1}& 19/0.3& 17/0.4&\textbf{4/0.2}&{9/0.3}&{14/0.3}\\
\midrule
  Ours (light, D2+SALAD, $\lambda=0.3$) & $\sim$244 MB & 18.7/0.1&{ 10.5/0.2}&14.6/0.3&4.8/0.2&{\ul 6.5/0.2}&11.0/0.2\\
   Ours (light, D2+MegaLoc, $\lambda=0.4$) & $\sim$244 MB & 21.2/0.1&10.7/0.2&\textbf{13.0}/0.3&4.7/0.2&6.6/0.2&11.2/0.2 \\
 Ours (light, D2+MixVPR, $\lambda=0.5$) & $\sim$244 MB & \textbf{15.6/0.1}&{ 10.5/0.2} &{\ul 13.5/0.3}&4.4/0.2& \textbf{6.3/0.2}&\textbf{10.0/0.2} \\
  Ours (heavy, D2+SALAD, $\lambda=0.3$)  & $\sim$287 MB & 18.9/0.1&{\ul 10.4/0.2}&14.6/0.3&4.5/0.2&6.6/0.2&11.0/0.2\\
  Ours (heavy, D2+MegaLoc, $\lambda=0.4$)  & $\sim$287 MB & 19.7/0.1&10.7/0.2&13.6/0.3&4.7/0.2&6.6/0.2&11.0/0.2\\
  Ours (heavy, D2+MixVPR, $\lambda=0.5$)  & $\sim$287 MB & 16.5/0.1&{ 10.5/0.2}&14.5/0.3&4.2/0.2&6.7/0.2& {\ul 10.5/0.2}\\
\bottomrule
\end{tabular}
}
\caption{\textbf{Cambridge Landmarks \cite{kendall2015posenet} Results.} We report median rotation (in degrees) and position errors (in cm). Other methods' statistics were gathered from~\cite{brachmann2023accelerated}. The best results are shown in \textbf{bold}, and the second best are \underline{underlined}. We outperform the vanilla system (which uses the local descriptors only), Active Search~\cite{active_search}, and hLoc~\cite{hloc} while using significantly less memory. Results for hLoc and Active Search were obtained from~\cite{brachmann2023accelerated}.}
\label{tab:results_cam}

\end{table*}

\subsection{Query time}
\label{subsec:querytime}
At query time, we first obtain the global descriptor $\globaldesc_{q}$ and the local descriptor $\localdesc_{iq}$  of the $i$-th keypoint in the \mbox{$q$-th} query image. We propose two variants of our method, balancing between memory footprint and accuracy. In the \textit{light} variant, the descriptor for the $i$-th keypoint for 2D-3D matching is computed as (similar to Equation~\ref{eq:codebook2}):
\begin{equation}
\label{eq:query1}
    \desc_{iq} = \lambda \localdesc_{iq} + (1-\lambda)\globaldesc_{q}.
\end{equation}
In the \textit{heavy} variant, we replace the query image's global descriptor  $\globaldesc_{q}$ with its nearest neighbor $\globaldesc_{k}$ among the database descriptors:
\begin{equation}
\label{eq:query2}
    \desc_{iq} = \lambda \localdesc_{iq} + (1-\lambda)\globaldesc_{k},
\end{equation}
where $k=\argmin_{k} \| \globaldesc_{k}-\globaldesc_{q} \|_2$. This additional step significantly improves accuracy at the cost of increased memory usage (by $25-30\%$; see Table~\ref{tab:memory_comparison}).

Finally, we search for the nearest neighbors of the combined descriptors in the codebook to obtain 2D-3D matching pairs.
\section{Experiments}
We first provide implementation details and hardware settings in Section~\ref{subsec:implementationdetails}. We evaluate our method on four popular datasets, which include outdoor scenes ranging from large buildings to city-level scales (Section~\ref{subsec:datasets}). We use the visual localization benchmark website\footnote{\url{www.visuallocalization.net/benchmark/}} for evaluation. The weights for the global descriptor networks~\cite{ali2023mixvpr, berton2023eigenplaces, Izquierdo_CVPR_2024_SALAD} and feature detectors~\cite{dusmanu2019d2, revaud2019r2d2} are obtained off-the-shelf, without any re-training or fine-tuning on the benchmark datasets.

\subsection{Implementation details}
\label{subsec:implementationdetails}
We performed ablation experiments (see Section~\ref{subsec:weight}) to determine the optimal parameter $\lambda$ for each global descriptor and reported their performance in Tables~\ref{tab:results_cam} and~\ref{tab:results_combined}. We note that $\lambda$ is determined only once during the codebook generation process and is adopted at query time. For each global descriptor method, we use the highest-performing variant as recommended by the authors.
We use 16-bit floating-point precision to store the codebook descriptors, and the FAISS~\cite{faiss} library to facilitate nearest neighbor lookup using a GPU. Final poses are estimated using RANSAC provided by PoseLib~\cite{PoseLib}. Our experiments were performed using a Linux system with a single NVIDIA A100 GPU and 100~GB of RAM. Note that the codebook training process is memory intensive and can consume up to 100~GB. However, this requirement drops to less than 10 GB at query time (see Table~\ref{tab:memory_comparison}).

\subsection{Datasets}
\label{subsec:datasets}

\begin{table*}[t]
\centering
\scriptsize
\resizebox{\textwidth}{!}{%
\setlength{\tabcolsep}{3pt}

\begin{tabular}{lccccccccccc}
\toprule
\multicolumn{1}{l}{}
  & &

\multicolumn{3}{c}{Aachen day/night v1.1} & \multicolumn{3}{c}{RobotCar Seasons v2} & \multicolumn{3}{c}{Extended CMU Seasons} \\
\cmidrule(l){3-5} 
\cmidrule(l){6-8} 
\cmidrule(l){9-11} 

& \multirow{-2}{*}{\begin{tabular}[c]{@{}c@{}}Memory \\ requirement $\downarrow$ \end{tabular}}  & 0.25m/2$^\circ$ & 0.5m/5$^\circ$ & 5m/10$^\circ$ & 0.25m/2$^\circ$ & 0.5m/5$^\circ$ & 5m/10$^\circ$ 
& 0.25m/2$^\circ$ & 0.5m/5$^\circ$ & 5m/10$^\circ$

& \multirow{-2}{*}{\begin{tabular}[c]{@{}c@{}}Average $\uparrow$ \\ (\%)\end{tabular}} \\

\midrule
  hLoc (SP+SG) \cite{hloc, sarlin2020superglue}&6.56 GB& \textbf{83.4} & \textbf{93.4} & \textbf{99.7} & \textbf{52.0} & \textbf{87.2} &  96.1 & \textbf{90.7} & \textbf{93.9} & { 96.0} & \textbf{88.0}\\

  GLACE \cite{GLACE2024CVPR}&13 MB &4.8 & 10.9 & 40.6  &-&-&-&-&-&-&-\\
  AS (SIFT) \cite{active_search}&-   &-&-&-&-&-&-&63.0 & 69.9 & 78.5&-\\
  R2D2~\cite{revaud2019r2d2} &0.75 GB&  68.9 & 76.7 & 85.7 & 32.4 & 51.8 & 59.4 & 55.9 & 60.1 & 68.6 & 62.2\\
  SIFT~\cite{DBLP:journals/ijcv/Lowe04} &0.75 GB&  56.0 & 60.0 & 65.6 & 24.9 & 39.2 & 43.6 & 34.6 & 38.6 & 45.4 & 45.3\\
  SuperPoint~\cite{detone2018superpoint} &1.5 GB&  67.4 & 77.7 & 85.6 & 31.0 & 51.4 & 61.1 & 60.1 & 65.0 & 72.6 & 63.5\\
  D2~\cite{dusmanu2019d2} &3 GB&  72.6 & 80.4 & 87.7 & 36.4 & 63.7 & 74.8& 78.8 & 83. & 89.8 & 74.2\\
  \midrule
 Ours (light, D2+SALAD, $\lambda=0.3$) \cite{dusmanu2019d2, Izquierdo_CVPR_2024_SALAD}&3 GB& 77.2 & 87.2 & 93.0 & 42.7 & 78.2 & 94.8 & 85.9 & 91.1 & 96.7 & 83.0 \\
 Ours (light, D2+MegaLoc, $\lambda=0.4$) \cite{dusmanu2019d2, berton2025megaloc}&3 GB&74.8 & 84.7 & 91.2 & 39.2 & 67.7 & 80.6 & 83.6 & 88.8 & 94.3 & 78.3\\
  Ours (light, D2+MixVPR, $\lambda=0.5$) \cite{dusmanu2019d2, ali2023mixvpr}&3 GB&72.4 & 81.4 & 89.0 & 37.4 & 64.7 & 76.6 & 83.0 & 88.1 & 93.9 & 76.3\\
  Ours (heavy, D2+SALAD, $\lambda=0.3$) \cite{dusmanu2019d2, Izquierdo_CVPR_2024_SALAD}&3.73 GB&{ 77.9} & { 88.7} & {\ul 95.5} & {\ul 47.0} & {\ul 87.0} & \textbf{99.6} & {\ul 88.5} & {\ul 93.6} & \textbf{98.0} & {\ul 86.2}\\
  Ours (heavy, D2+MegaLoc, $\lambda=0.4$) \cite{dusmanu2019d2, berton2025megaloc}&3.73 GB&{\ul79.8} & {\ul88.8} & 95.4 & 46.5 & 84.8 & {\ul98.5} & 88.0 & 93.0 & {\ul97.7} & 85.8\\
  Ours (heavy, D2+MixVPR, $\lambda=0.5$) \cite{dusmanu2019d2, ali2023mixvpr}&3.13 GB&77.4 & 86.3 & 91.4 & 42.0 & 73.9 & 89.2 & 87.5 & 92.6 & 97.4 & 82.0\\

\bottomrule
\end{tabular}
}
\caption{\textbf{Visual Localization benchmark Results.} We report the percentage of query images successfully localized under different thresholds. The best results are shown in \textbf{bold}, and the second best are \underline{underlined}. Our global+local method on average improves $12 \%$ over the vanilla systems. We further narrow the performance gap to hierarchical methods to just $1.8 \%$ on average, compared to the best performing local-descriptor-only technique which performs $14 \%$ worse than hierarchical methods. The statistics for the memory requirements are averaged over three datasets. Results for hLoc~\cite{hloc} were obtained from the official benchmark website under the following IDs: ``Hierarchical Localization -- SuperPoint + SuperGlue'' and ``Hierarchical Localization -- SuperPoint + SuperGlue (single-camera)''.}
\label{tab:results_combined}

\end{table*}
\begin{table}[th]

  \centering
  \footnotesize
  %\scriptsize
  \setlength{\tabcolsep}{2.75pt}
\begin{tabular}{lccc}
\toprule
\multicolumn{1}{l}{} & \multicolumn{3}{c}{Aachen / CMU / RobotCar (GB)}   \\
\cmidrule(l){2-4} 
\multicolumn{1}{l}{} & hLoc~\cite{hloc} & ours (light) & ours (heavy)\\

\midrule

  Codebook &-& 2.0 / 2.0 / 5.0 & 2.0 / 2.0 / 5.0\\
  Database images & 5.0 / 4.0 / 6.0 & - & - \\
  Database image descriptors &0.4 / 0.4 / 1.4 & - & 0.4 / 0.4 / 1.4\\
  Pixel-to-point mappings &0.5 / 1.0 / 2.0 & - & -\\\midrule
  \textbf{Total} & 5.9 / 5.4 / 8.4 & 2.0 / 2.0 / 5.0 & 2.4 / 2.4 / 6.4\\
\bottomrule
\end{tabular}
\caption{\textbf{Memory requirement comparison.} We estimate the memory requirements for each method for three different datasets. 
We omitted the networks' weights, and the 3D coordinates of the map, as all three methods require this information. 
All statistics for our method were gathered using D2~\cite{dusmanu2019d2} and SALAD~\cite{Izquierdo_CVPR_2024_SALAD}. 
}
\label{tab:memory_comparison}

\end{table}
\begin{table}[th]
  \centering
  \footnotesize
  % \resizebox{\columnwidth}{!}{%
  \begin{tabular}{lcccc}
    \toprule
    \multicolumn{1}{l}{} & hloc~\cite{hloc} & ours (heavy) & ours (light)\\
    \midrule
    Aachen & 0.19 & \textbf{0.30} & \textbf{0.30} \\
    RobotCar & 0.32 & 0.56 & \textbf{0.57} \\
    CMU & 0.40 & 0.58 & \textbf{0.69} \\
    \midrule
    \textbf{Average} & 0.3 & 0.48 & \textbf{0.52} \\
    \bottomrule
  \end{tabular}%
  % }
  \caption{\textbf{Frames per second.}  We report the frame rate for hLoc~\cite{hloc} and our method using D2~\cite{dusmanu2019d2} and SALAD~\cite{Izquierdo_CVPR_2024_SALAD}. 
  We excluded the processing of the local and global descriptors of the query images since both methods perform these steps. Note that our light version is faster but not as accurate as our heavy version.}
  \label{tab:query_time_comparison}
\end{table}

The \textbf{Cambridge Landmarks Dataset}~\cite{DBLP:conf/iccv/KendallGC15} was collected at the University of Cambridge using a Google LG \mbox{Nexus 5} smartphone in large-scale outdoor urban environments. HD videos were recorded at 2 Hz to extract RGB frames. The dataset features typical urban challenges such as moving pedestrians and vehicles, and captures different lighting and weather conditions over time. Training and test data were obtained from disjoint walking trajectories, increasing the difficulty of the localization task. Ground-truth camera poses and 3D models were generated using VisualSfM~\cite{wu2013towards} and are provided with the dataset. 

% aachen
The \textbf{Aachen Day-Night v1.1 Dataset}~\cite{zhang2021reference} enhances the original Aachen dataset~\cite{sattler2018benchmarking} with new sequences to construct a comprehensive 3D model of the historic inner city of Aachen, Germany, using COLMAP~\cite{schoenberger2016sfm}. Training images were captured during the daytime, while the test set includes nighttime images processed using HDR software to improve illumination.

\textbf{RobotCar Seasons v2}~\cite{Maddern2017IJRR} encompasses 20 million images collected over a one year period in a variety of weather conditions using an autonomous car equipped with six cameras, covering over 1000 km in Oxford, UK. For benchmarking, 49 different sub-models (each covering different locations) were reconstructed using high-quality images captured under overcast conditions. The test set includes images taken under a broader range of weather conditions. We evaluated algorithms against a global model containing the entire map rather than individual sub-models.

% extended cmu
\textbf{Extended CMU Seasons}~\cite{sattler2018benchmarking} was captured at the Carnegie Mellon University over 12 months under different weather conditions. A vehicle was equipped with two cameras and completed 16 traverses following an 8.5 km route through central and suburban Pittsburgh. A 3D model of the scene was constructed using images taken under good weather conditions (sunny with no foliage). This 3D model was used to generate reference poses for all remaining dataset images.

\begin{figure}[t]
    \centering
    \includegraphics[width=1.0\columnwidth]{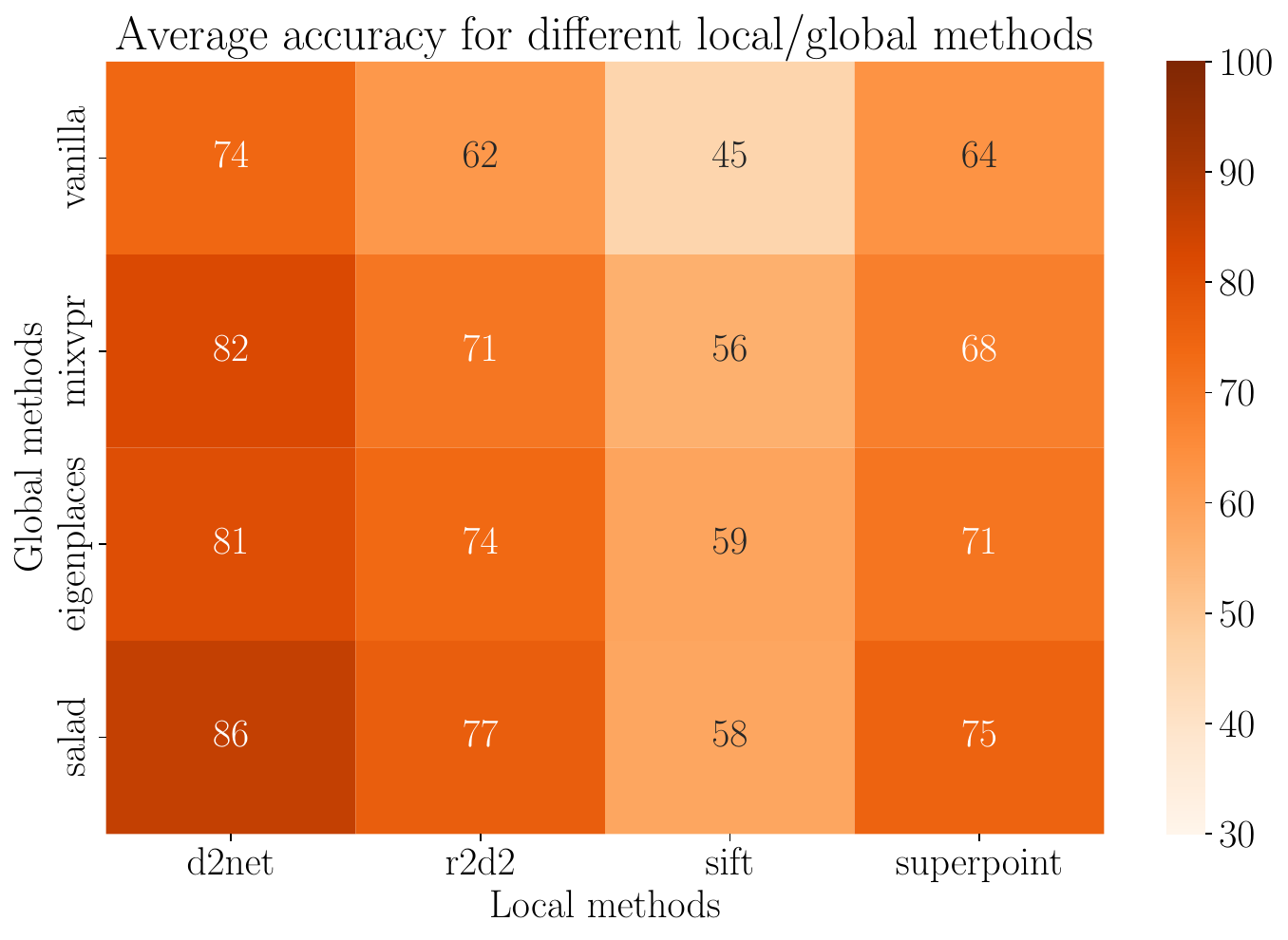}
    \caption{\textbf{Performance with different local/global methods.} We present a complete overview of different local and global features on Aachen Day/Night v1.1~\cite{zhang2021reference}, RobotCar Seasons v2~\cite{Maddern2017IJRR} and Extended CMU seasons~\cite{sattler2018benchmarking}. We report the average percentage of successfully localized images under different thresholds for all three datasets. We consistently observe a significant boost when using global descriptors compared to using only local descriptors across all four different local descriptors~(\textit{vanilla, top row}). }
    \label{fig:heatmap}
\end{figure}

\section{Evaluation}

\subsection{Quantitative comparison}
Our method consistently outperforms the local-descriptor-only codebooks across all datasets (Tables~\ref{tab:results_cam} and~\ref{tab:results_combined}). For example, the average median translation error is reduced from $12.1$ cm to $10.0$ cm when fusing D2 with MixVPR on the Cambridge Landmarks~\cite{kendall2015posenet} dataset (Table~\ref{tab:results_cam}). On larger maps (Aachen Day/Night v1.1~\cite{zhang2021reference}, RobotCar Seasons v2~\cite{Maddern2017IJRR} and Extended CMU seasons~\cite{sattler2018benchmarking}), our method improves the percentage of successfully localized test images over the D2 local descriptor~\cite{dusmanu2019d2} from $74.2\%$ to $86.2\%$ on average, depending on the variant used (Table~\ref{tab:results_combined}). We refer the reader to Figure~\ref{fig:heatmap} for a complete overview of evaluation results averaged over these three large-scale datasets for different combinations of global and local descriptors.

% vs hloc
Our method further narrows the performance gap between direct matching algorithms~\cite{DBLP:conf/eccv/LiSH10, active_search, sattler2011fast} and hierarchical algorithms~\cite{hloc, peng2021megloc}, while retaining the appealing computational complexity of direct matching methods. On the Cambridge Landmarks dataset~\cite{kendall2015posenet}, we achieve an improvement of almost 1 cm ($7.7\%$) in average median translation error over hLoc~\cite{hloc} while using only around $5\%$ of the memory required (Table~\ref{tab:results_cam}). On larger maps with substantial perceptual aliasing, our method performs only $1.8\%$ worse than hLoc (Table~\ref{tab:results_combined}) while using only $57\%$ of the memory footprint (Table~\ref{tab:memory_comparison}) and being $1.6$ times faster on average (Table~\ref{tab:query_time_comparison}).

\subsection{Computational complexity}
Table~\ref{tab:memory_comparison} measures the memory requirements of the required components for each method, including the codebook, the database images and their descriptors, and the mapping from database image pixels to point cloud coordinates. Since all methods under consideration require the point cloud coordinates and the weights of the deep networks, we exclude their memory footprints from the values reported in Table~\ref{tab:memory_comparison}. Depending on the dataset and whether our light or heavy variant is used, the memory requirements are roughly half of that of hLoc~\cite{hloc}. 

We report our frame rate in Table~\ref{tab:query_time_comparison}. By performing direct matching against 3D points, we eliminate the costly feature detection step for database images. While caching these database features could accelerate hierarchical algorithms, it would demand approximately 30-50 times more memory. For example, storing only the D2~\cite{dusmanu2019d2} local features consumed 158.4 GB for the Aachen dataset and 223.3 GB for the RobotCar dataset. Given the memory settings outlined in Table~\ref{tab:memory_comparison}, our method achieves a speedup of approximately 1.6x compared to hLoc~\cite{hloc}.

\begin{figure}[t]
    \centering
    \includegraphics[width=\columnwidth]{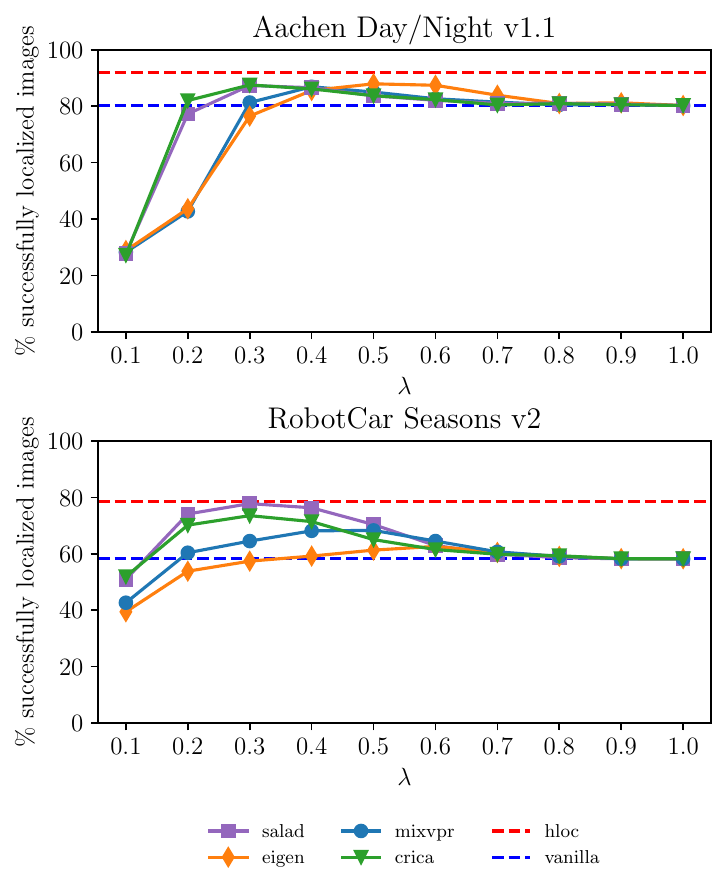}
    \caption{\textbf{Sensitivity of $\lambda$.} This figure illustrates the impact of varying the $\lambda$ parameter on the performance of our method. Note that $\lambda~=~1.0$ (blue dotted lines) corresponds to the vanilla codebook using only local descriptors. With the D2+SALAD descriptor and $\lambda~=~0.3$, our method achieves performance close to that of hloc's on both Aachen Day/Night v1.1~\cite{zhang2021reference} (top) and RobotCar Seasons v2~\cite{Maddern2017IJRR} (bottom) datasets.}
    \label{fig:ablation}
    \vspace*{-.35cm}
\end{figure}

\begin{figure}[t]
    \centering
    \includegraphics[width=\columnwidth]{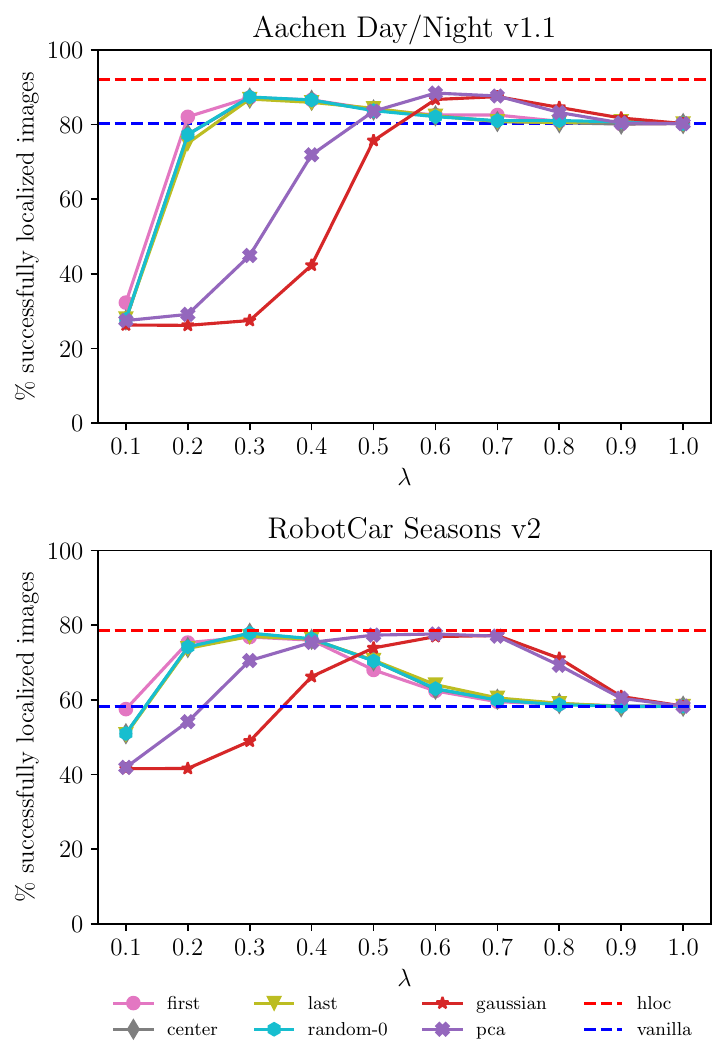}
    \caption{\textbf{Performance with different methods to truncate global descriptors.} This figure compares the performance of various methods for truncating global descriptors. All methods yield comparable performance, with \textit{random-0} and \textit{pca} slightly outperforming the others. We use \textit{random-0} for all of our experiments due to its simplicity.}
    \label{fig:ablation_order}
\end{figure}

\section{Ablation studies} \label{sec:ablation}
\subsection{Hyperparameter \texorpdfstring{$\lambda$}{lambda}}
\label{subsec:weight}
We conducted extensive experiments to examine the sensitivity of the parameter $\lambda$ on two datasets: Aachen Day/Night v1.1~\cite{zhang2021reference} and RobotCar Seasons v2~\cite{Maddern2017IJRR}. Using the \textit{heavy} variant of our system, we tested ten different $\lambda$ values ranging from $0.1$ to $1.0$. Note that $\lambda=1.0$ corresponds to the vanilla codebook. 

Figure~\ref{fig:ablation} shows the percentage of successfully localized images for each $\lambda$ value. We found that the \mbox{optimal $\lambda$} for SALAD~\cite{Izquierdo_CVPR_2024_SALAD} and CRICA~\cite{lu2024cricavpr} lies between $0.3$ and $0.4$. This setting results in a codebook that significantly enhances performance compared to the default $\lambda=0.5$, achieving performance levels very close to hierarchical algorithms~\cite{hloc} (only $4.6\%$ and $0.7\%$ performance reduction on Aachen and RobotCar, respectively) while using $43\%$ less memory (Table~\ref{tab:memory_comparison}) and being $1.6$ times faster (Table~\ref{tab:query_time_comparison}). Furthermore, we note that our system performs well for a wide range of $\lambda$, consistently improving upon the vanilla system that only uses local descriptors for $0.2\leq\lambda\leq0.7$.

\subsection{Global descriptor truncation}
\label{subsec:descriptortruncation}
We tested different methods to truncate the global descriptors:
\begin{itemize}
  \item \textit{gaussian:} Gaussian random projection \cite{bingham2001random},
  \item \textit{pca:} PCA projection \cite{pca},
  \item \textit{random-0:} random order with seed $0$,
  \item \textit{first:} keep only the first $m$ entries,
  \item \textit{center:} keep only the $m$ entries around the middle index,
  \item \textit{last:} keep only the last $m$ entries.
\end{itemize}

Using the D2+SALAD variant, we again varied $\lambda$ from $0.1$ to $1.0$. Figure~\ref{fig:ablation_order} shows that most truncation methods perform comparably, with \textit{random-0} and \textit{pca} performing slightly better than the other techniques. Therefore, we use \textit{random-0} for all of our experiments due to its simplicity.

\section{Conclusion}
We introduce a simple technique to enhance 2D-3D search in direct matching visual localization. Through extensive evaluation on four real-world datasets, we demonstrate that our method significantly improves the performance of a brute-force baseline system with minimal memory overhead. Our ablation studies show that our approach performs comparably to hierarchical methods while using $43\%$ less memory and running $1.6$ times faster, making our method particularly appealing for robotic systems with limited onboard memory.

The primary limitation of our method is that it can only disambiguate non-co-visible points in the database map. We recommend that future research focuses on identifying potential clues for resolving ambiguities among co-visible points. We hope our work will continue to spark the community's interest in 2D-3D matching systems due to their lower memory consumption and high potential performance when combined with disambiguation techniques.

\section*{Acknowledgements}This research was partially supported by funding from ARC Laureate Fellowship FL210100156 to MM, ARC DECRA Fellowship DE240100149 to TF, and the QUT Centre for Robotics.
{
    \small
    \bibliographystyle{ieeenat_fullname}
    \bibliography{main}

\begin{thebibliography}{56}
\providecommand{\natexlab}[1]{#1}
\providecommand{\url}[1]{\texttt{#1}}
\expandafter\ifx\csname urlstyle\endcsname\relax
  \providecommand{\doi}[1]{doi: #1}\else
  \providecommand{\doi}{doi: \begingroup \urlstyle{rm}\Url}\fi

\bibitem[Ali-Bey et~al.(2023)Ali-Bey, Chaib-Draa, and Giguere]{ali2023mixvpr}
Amar Ali-Bey, Brahim Chaib-Draa, and Philippe Giguere.
\newblock Mixvpr: Feature mixing for visual place recognition.
\newblock In \emph{IEEE Winter Conf. Applicat. Comput. Vis.}, pages 2998--3007,
  2023.

\bibitem[Arandjelovic et~al.(2016)Arandjelovic, Gron{\'{a}}t, Torii, Pajdla,
  and Sivic]{DBLP:conf/cvpr/ArandjelovicGTP16}
Relja Arandjelovic, Petr Gron{\'{a}}t, Akihiko Torii, Tom{\'{a}}s Pajdla, and
  Josef Sivic.
\newblock Netvlad: {CNN} architecture for weakly supervised place recognition.
\newblock In \emph{IEEE Conf. Comput. Vis. Pattern Recog.}, pages 5297--5307,
  2016.

\bibitem[Arnold et~al.(2022)Arnold, Wynn, Vicente, Garcia-Hernando, Monszpart,
  Prisacariu, Turmukhambetov, and Brachmann]{arnold2022map}
Eduardo Arnold, Jamie Wynn, Sara Vicente, Guillermo Garcia-Hernando, {\'A}ron
  Monszpart, Victor Prisacariu, Daniyar Turmukhambetov, and Eric Brachmann.
\newblock Map-free visual relocalization: Metric pose relative to a single
  image.
\newblock In \emph{Eur. Conf. Comput. Vis.}, pages 690--708, 2022.

\bibitem[Badino et~al.(2011)Badino, Huber, and Kanade]{cmu_dataset}
Hernan Badino, Daniel Huber, and Takeo Kanade.
\newblock {The CMU Visual Localization Data Set}.
\newblock \url{http://3dvis.ri.cmu.edu/data-sets/localization}, 2011.

\bibitem[Bay et~al.(2006)Bay, Tuytelaars, and Gool]{DBLP:conf/eccv/BayTG06}
Herbert Bay, Tinne Tuytelaars, and Luc~Van Gool.
\newblock {SURF:} speeded up robust features.
\newblock In \emph{Eur. Conf. Comput. Vis.}, pages 404--417, 2006.

\bibitem[Berton and Masone(2025)]{berton2025megaloc}
Gabriele Berton and Carlo Masone.
\newblock Megaloc: One retrieval to place them all.
\newblock In \emph{IEEE Conf. Comput. Vis. Pattern Recog. Worksh.}, pages
  2861--2867, 2025.

\bibitem[Berton et~al.(2023)Berton, Trivigno, Caputo, and
  Masone]{berton2023eigenplaces}
Gabriele Berton, Gabriele Trivigno, Barbara Caputo, and Carlo Masone.
\newblock Eigenplaces: Training viewpoint robust models for visual place
  recognition.
\newblock In \emph{IEEE Int. Conf. Comput. Vis.}, pages 11080--11090, 2023.

\bibitem[Bingham and Mannila(2001)]{bingham2001random}
Ella Bingham and Heikki Mannila.
\newblock Random projection in dimensionality reduction: applications to image
  and text data.
\newblock In \emph{ACM SIGKDD}, pages 245--250, 2001.

\bibitem[Brachmann and Rother(2018)]{brachmann2018learning}
Eric Brachmann and Carsten Rother.
\newblock Learning less is more-6d camera localization via 3d surface
  regression.
\newblock In \emph{IEEE Conf. Comput. Vis. Pattern Recog.}, pages 4654--4662,
  2018.

\bibitem[Brachmann and Rother(2022)]{DBLP:journals/pami/BrachmannR22}
Eric Brachmann and Carsten Rother.
\newblock Visual camera re-localization from {RGB} and {RGB-D} images using
  {DSAC}.
\newblock \emph{IEEE Trans. Pattern Anal. Mach. Intell.}, 44\penalty0
  (9):\penalty0 5847--5865, 2022.

\bibitem[Brachmann et~al.(2023)Brachmann, Cavallari, and
  Prisacariu]{brachmann2023accelerated}
Eric Brachmann, Tommaso Cavallari, and Victor~Adrian Prisacariu.
\newblock Accelerated coordinate encoding: Learning to relocalize in minutes
  using rgb and poses.
\newblock In \emph{IEEE Conf. Comput. Vis. Pattern Recog.}, pages 5044--5053,
  2023.

\bibitem[Cao et~al.(2020)Cao, Araujo, and Sim]{cao2020unifying}
Bingyi Cao, Andre Araujo, and Jack Sim.
\newblock Unifying deep local and global features for image search.
\newblock In \emph{Eur. Conf. Comput. Vis.}, pages 726--743, 2020.

\bibitem[Chen et~al.(2024)Chen, Cavallari, Prisacariu, and
  Brachmann]{chen2024map}
Shuai Chen, Tommaso Cavallari, Victor~Adrian Prisacariu, and Eric Brachmann.
\newblock Map-relative pose regression for visual re-localization.
\newblock In \emph{IEEE Conf. Comput. Vis. Pattern Recog.}, pages 20665--20674,
  2024.

\bibitem[DeTone et~al.(2018)DeTone, Malisiewicz, and
  Rabinovich]{detone2018superpoint}
Daniel DeTone, Tomasz Malisiewicz, and Andrew Rabinovich.
\newblock Superpoint: Self-supervised interest point detection and description.
\newblock In \emph{IEEE Conf. Comput. Vis. Pattern Recog. Worksh.}, pages
  224--236, 2018.

\bibitem[Dusmanu et~al.(2019)Dusmanu, Rocco, Pajdla, Pollefeys, Sivic, Torii,
  and Sattler]{dusmanu2019d2}
Mihai Dusmanu, Ignacio Rocco, Tomas Pajdla, Marc Pollefeys, Josef Sivic,
  Akihiko Torii, and Torsten Sattler.
\newblock D2-net: A trainable cnn for joint description and detection of local
  features.
\newblock In \emph{IEEE Conf. Comput. Vis. Pattern Recog.}, pages 8092--8101,
  2019.

\bibitem[Halko et~al.(2011)Halko, Martinsson, and Tropp]{pca}
Nathan Halko, Per-Gunnar Martinsson, and Joel~A Tropp.
\newblock Finding structure with randomness: Probabilistic algorithms for
  constructing approximate matrix decompositions.
\newblock \emph{SIAM review}, 53\penalty0 (2):\penalty0 217--288, 2011.

\bibitem[Hausler et~al.(2021)Hausler, Garg, Xu, Milford, and
  Fischer]{hausler2021patch}
Stephen Hausler, Sourav Garg, Ming Xu, Michael Milford, and Tobias Fischer.
\newblock Patch-netvlad: Multi-scale fusion of locally-global descriptors for
  place recognition.
\newblock In \emph{IEEE Conf. Comput. Vis. Pattern Recog.}, pages 14141--14152,
  2021.

\bibitem[Irschara et~al.(2009)Irschara, Zach, Frahm, and
  Bischof]{DBLP:conf/cvpr/IrscharaZFB09}
Arnold Irschara, Christopher Zach, Jan{-}Michael Frahm, and Horst Bischof.
\newblock From structure-from-motion point clouds to fast location recognition.
\newblock In \emph{IEEE Conf. Comput. Vis. Pattern Recog.}, pages 2599--2606,
  2009.

\bibitem[Izquierdo and Civera(2024)]{Izquierdo_CVPR_2024_SALAD}
Sergio Izquierdo and Javier Civera.
\newblock Optimal transport aggregation for visual place recognition.
\newblock In \emph{IEEE Conf. Comput. Vis. Pattern Recog.}, pages 17658--17668,
  2024.

\bibitem[Johnson et~al.(2019)Johnson, Douze, and J{\'e}gou]{faiss}
Jeff Johnson, Matthijs Douze, and Herv{\'e} J{\'e}gou.
\newblock Billion-scale similarity search with {GPUs}.
\newblock \emph{IEEE Trans. Big Data}, 7\penalty0 (3):\penalty0 535--547, 2019.

\bibitem[Kendall et~al.(2015{\natexlab{a}})Kendall, Grimes, and
  Cipolla]{DBLP:conf/iccv/KendallGC15}
Alex Kendall, Matthew Grimes, and Roberto Cipolla.
\newblock Posenet: {A} convolutional network for real-time 6-dof camera
  relocalization.
\newblock In \emph{IEEE Int. Conf. Comput. Vis.}, pages 2938--2946,
  2015{\natexlab{a}}.

\bibitem[Kendall et~al.(2015{\natexlab{b}})Kendall, Grimes, and
  Cipolla]{kendall2015posenet}
Alex Kendall, Matthew Grimes, and Roberto Cipolla.
\newblock Posenet: A convolutional network for real-time 6-dof camera
  relocalization.
\newblock In \emph{IEEE Int. Conf. Comput. Vis.}, pages 2938--2946,
  2015{\natexlab{b}}.

\bibitem[Kim et~al.(2015)Kim, Dunn, and Frahm]{kim2015predicting}
Hyo~Jin Kim, Enrique Dunn, and Jan-Michael Frahm.
\newblock Predicting good features for image geo-localization using per-bundle
  vlad.
\newblock In \emph{IEEE Int. Conf. Comput. Vis.}, pages 1170--1178, 2015.

\bibitem[Larsson(2020)]{PoseLib}
Viktor Larsson.
\newblock {PoseLib - Minimal Solvers for Camera Pose Estimation}, 2020.

\bibitem[Laskar et~al.(2024)Laskar, Melekhov, Benbihi, Wang, and
  Kannala]{laskar2024differentiable}
Zakaria Laskar, Iaroslav Melekhov, Assia Benbihi, Shuzhe Wang, and Juho
  Kannala.
\newblock Differentiable product quantization for memory efficient camera
  relocalization.
\newblock In \emph{Eur. Conf. Comput. Vis.}, pages 470--489, 2024.

\bibitem[Li et~al.(2010)Li, Snavely, and Huttenlocher]{DBLP:conf/eccv/LiSH10}
Yunpeng Li, Noah Snavely, and Daniel~P. Huttenlocher.
\newblock Location recognition using prioritized feature matching.
\newblock In \emph{Eur. Conf. Comput. Vis.}, pages 791--804, 2010.

\bibitem[Loquercio et~al.(2017)Loquercio, Dymczyk, Zeisl, Lynen, Gilitschenski,
  and Siegwart]{loquercio2017efficient}
Antonio Loquercio, Marcin Dymczyk, Bernhard Zeisl, Simon Lynen, Igor
  Gilitschenski, and Roland Siegwart.
\newblock Efficient descriptor learning for large scale localization.
\newblock In \emph{IEEE Int. Conf. Robot. Autom.}, pages 3170--3177, 2017.

\bibitem[Lowe(2004)]{DBLP:journals/ijcv/Lowe04}
David~G. Lowe.
\newblock Distinctive image features from scale-invariant keypoints.
\newblock \emph{Int. J. Comput. Vis.}, 60\penalty0 (2):\penalty0 91--110, 2004.

\bibitem[Lowry et~al.(2016)Lowry, S{\"{u}}nderhauf, Newman, Leonard, Cox,
  Corke, and Milford]{DBLP:journals/trob/LowryS0LCCM16}
Stephanie~M. Lowry, Niko S{\"{u}}nderhauf, Paul Newman, John~J. Leonard,
  David~D. Cox, Peter~I. Corke, and Michael~J. Milford.
\newblock Visual place recognition: {A} survey.
\newblock \emph{IEEE Trans. Robot.}, 32\penalty0 (1):\penalty0 1--19, 2016.

\bibitem[Lu et~al.(2024)Lu, Lan, Zhang, Jiang, Wang, and Yuan]{lu2024cricavpr}
Feng Lu, Xiangyuan Lan, Lijun Zhang, Dongmei Jiang, Yaowei Wang, and Chun Yuan.
\newblock {CricaVPR: Cross-image} correlation-aware representation learning for
  visual place recognition.
\newblock In \emph{IEEE Conf. Comput. Vis. Pattern Recog.}, pages 16772--16782,
  2024.

\bibitem[Maddern et~al.(2017)Maddern, Pascoe, Linegar, and
  Newman]{Maddern2017IJRR}
Will Maddern, Geoffrey Pascoe, Chris Linegar, and Paul Newman.
\newblock {1 Year, 1000km: The Oxford RobotCar Dataset}.
\newblock \emph{Int. J. Robot. Res.}, 36\penalty0 (1):\penalty0 3--15, 2017.

\bibitem[Masone and Caputo(2021)]{masone2021survey}
Carlo Masone and Barbara Caputo.
\newblock A survey on deep visual place recognition.
\newblock \emph{IEEE Access}, 9:\penalty0 19516--19547, 2021.

\bibitem[Mohedano et~al.(2016)Mohedano, McGuinness, O'Connor, Salvador,
  Marqu{\'{e}}s, and Gir{\'{o}}{-}i{-}Nieto]{DBLP:conf/mir/MohedanoMOSMN16}
Eva Mohedano, Kevin McGuinness, Noel~E. O'Connor, Amaia Salvador, Ferran
  Marqu{\'{e}}s, and Xavier Gir{\'{o}}{-}i{-}Nieto.
\newblock Bags of local convolutional features for scalable instance search.
\newblock In \emph{Int. Conf. Multimedia Retrieval}, pages 327--331, 2016.

\bibitem[Nguyen et~al.(2024)Nguyen, Fontan, Milford, and Fischer]{focustune}
Son~Tung Nguyen, Alejandro Fontan, Michael Milford, and Tobias Fischer.
\newblock Focustune: Tuning visual localization through focus-guided sampling.
\newblock In \emph{IEEE Winter Conf. Applicat. Comput. Vis.}, pages 3606--3615,
  2024.

\bibitem[Peng et~al.(2021)Peng, He, Zhang, Yan, Wang, Zhu, and
  Liu]{peng2021megloc}
Shuxue Peng, Zihang He, Haotian Zhang, Ran Yan, Chuting Wang, Qingtian Zhu, and
  Xiao Liu.
\newblock Megloc: A robust and accurate visual localization pipeline.
\newblock \emph{arXiv preprint arXiv:2111.13063}, 2021.

\bibitem[Phan et~al.(2022)Phan, Nguyen, Warrier, and Gupta]{phan2022patch}
Lam Phan, Hiep Thi~Hong Nguyen, Harikrishna Warrier, and Yogesh Gupta.
\newblock Patch embedding as local features: Unifying deep local and global
  features via vision transformer for image retrieval.
\newblock In \emph{ACCV}, pages 2527--2544, 2022.

\bibitem[Revaud et~al.(2019)Revaud, De~Souza, Humenberger, and
  Weinzaepfel]{revaud2019r2d2}
Jerome Revaud, Cesar De~Souza, Martin Humenberger, and Philippe Weinzaepfel.
\newblock R2d2: Reliable and repeatable detector and descriptor.
\newblock \emph{Adv. Neural Inform. Process. Syst.}, 32, 2019.

\bibitem[Sarlin et~al.(2019)Sarlin, Cadena, Siegwart, and Dymczyk]{hloc}
Paul{-}Edouard Sarlin, Cesar Cadena, Roland Siegwart, and Marcin Dymczyk.
\newblock From coarse to fine: Robust hierarchical localization at large scale.
\newblock In \emph{IEEE Conf. Comput. Vis. Pattern Recog.}, pages 12716--12725,
  2019.

\bibitem[Sarlin et~al.(2020)Sarlin, DeTone, Malisiewicz, and
  Rabinovich]{sarlin2020superglue}
Paul-Edouard Sarlin, Daniel DeTone, Tomasz Malisiewicz, and Andrew Rabinovich.
\newblock Superglue: Learning feature matching with graph neural networks.
\newblock In \emph{IEEE Conf. Comput. Vis. Pattern Recog.}, pages 4938--4947,
  2020.

\bibitem[Sattler et~al.(2011)Sattler, Leibe, and Kobbelt]{sattler2011fast}
Torsten Sattler, Bastian Leibe, and Leif Kobbelt.
\newblock Fast image-based localization using direct 2d-to-3d matching.
\newblock In \emph{IEEE Int. Conf. Comput. Vis.}, pages 667--674, 2011.

\bibitem[Sattler et~al.(2012{\natexlab{a}})Sattler, Leibe, and
  Kobbelt]{active_search}
Torsten Sattler, Bastian Leibe, and Leif Kobbelt.
\newblock Improving image-based localization by active correspondence search.
\newblock In \emph{Eur. Conf. Comput. Vis.}, pages 752--765,
  2012{\natexlab{a}}.

\bibitem[Sattler et~al.(2012{\natexlab{b}})Sattler, Weyand, Leibe, and
  Kobbelt]{sattler2012image}
Torsten Sattler, Tobias Weyand, Bastian Leibe, and Leif Kobbelt.
\newblock Image retrieval for image-based localization revisited.
\newblock In \emph{Brit. Mach. Vis. Conf.}, 2012{\natexlab{b}}.

\bibitem[Sattler et~al.(2016)Sattler, Leibe, and Kobbelt]{sattler2016efficient}
Torsten Sattler, Bastian Leibe, and Leif Kobbelt.
\newblock Efficient \& effective prioritized matching for large-scale
  image-based localization.
\newblock \emph{IEEE Trans. Pattern Anal. Mach. Intell.}, 39\penalty0
  (9):\penalty0 1744--1756, 2016.

\bibitem[Sattler et~al.(2018)Sattler, Maddern, Toft, Torii, Hammarstrand,
  Stenborg, Safari, Okutomi, Pollefeys, Sivic, et~al.]{sattler2018benchmarking}
Torsten Sattler, Will Maddern, Carl Toft, Akihiko Torii, Lars Hammarstrand,
  Erik Stenborg, Daniel Safari, Masatoshi Okutomi, Marc Pollefeys, Josef Sivic,
  et~al.
\newblock Benchmarking 6dof outdoor visual localization in changing conditions.
\newblock In \emph{IEEE Conf. Comput. Vis. Pattern Recog.}, pages 8601--8610,
  2018.

\bibitem[Sch\"{o}nberger and Frahm(2016)]{schoenberger2016sfm}
Johannes~Lutz Sch\"{o}nberger and Jan-Michael Frahm.
\newblock Structure-from-motion revisited.
\newblock In \emph{IEEE Conf. Comput. Vis. Pattern Recog.}, pages 4104--4113,
  2016.

\bibitem[Schubert et~al.(2023)Schubert, Neubert, Garg, Milford, and
  Fischer]{schubert2023visual}
Stefan Schubert, Peer Neubert, Sourav Garg, Michael Milford, and Tobias
  Fischer.
\newblock Visual place recognition: A tutorial.
\newblock \emph{{IEEE} Robot. Automat. Mag.}, 31\penalty0 (3):\penalty0
  139--153, 2023.

\bibitem[Shavit et~al.(2021)Shavit, Ferens, and
  Keller]{DBLP:conf/iccv/ShavitFK21}
Yoli Shavit, Ron Ferens, and Yosi Keller.
\newblock Learning multi-scene absolute pose regression with transformers.
\newblock In \emph{IEEE Int. Conf. Comput. Vis.}, pages 2713--2722, 2021.

\bibitem[Shotton et~al.(2013)Shotton, Glocker, Zach, Izadi, Criminisi, and
  Fitzgibbon]{DBLP:conf/cvpr/ShottonGZICF13}
Jamie Shotton, Ben Glocker, Christopher Zach, Shahram Izadi, Antonio Criminisi,
  and Andrew~W. Fitzgibbon.
\newblock Scene coordinate regression forests for camera relocalization in
  {RGB-D} images.
\newblock In \emph{IEEE Conf. Comput. Vis. Pattern Recog.}, pages 2930--2937,
  2013.

\bibitem[Tolias et~al.(2013)Tolias, Avrithis, and
  J{\'e}gou]{tolias2013aggregate}
Giorgos Tolias, Yannis Avrithis, and Herv{\'e} J{\'e}gou.
\newblock To aggregate or not to aggregate: Selective match kernels for image
  search.
\newblock In \emph{IEEE Int. Conf. Comput. Vis.}, pages 1401--1408, 2013.

\bibitem[Wang et~al.(2024{\natexlab{a}})Wang, Jiang, Galliani, Vogel, and
  Pollefeys]{GLACE2024CVPR}
Fangjinhua Wang, Xudong Jiang, Silvano Galliani, Christoph Vogel, and Marc
  Pollefeys.
\newblock {GLACE: Global Local Accelerated Coordinate Encoding}.
\newblock In \emph{IEEE Conf. Comput. Vis. Pattern Recog.}, pages 21562--21571,
  2024{\natexlab{a}}.

\bibitem[Wang(2024)]{wang2024mad}
Qiang Wang.
\newblock {MAD-DR: Map} compression for visual localization with matchness
  aware descriptor dimension reduction.
\newblock In \emph{Eur. Conf. Comput. Vis.}, pages 261--278, 2024.

\bibitem[Wang et~al.(2024{\natexlab{b}})Wang, Leroy, Cabon, Chidlovskii, and
  Revaud]{duster}
Shuzhe Wang, Vincent Leroy, Yohann Cabon, Boris Chidlovskii, and Jerome Revaud.
\newblock Dust3r: Geometric 3d vision made easy.
\newblock In \emph{IEEE Conf. Comput. Vis. Pattern Recog.}, pages 20697--20709,
  2024{\natexlab{b}}.

\bibitem[Wu(2013)]{wu2013towards}
Changchang Wu.
\newblock Towards linear-time incremental structure from motion.
\newblock In \emph{Int. Conf. 3D Vision}, pages 127--134, 2013.

\bibitem[Wu et~al.(2023)Wu, Wang, Zhou, Lu, and Li]{wu2023asymmetric}
Hui Wu, Min Wang, Wengang Zhou, Zhenbo Lu, and Houqiang Li.
\newblock Asymmetric feature fusion for image retrieval.
\newblock In \emph{IEEE Conf. Comput. Vis. Pattern Recog.}, pages 11082--11092,
  2023.

\bibitem[Yang et~al.(2021)Yang, He, Fan, Shi, Xue, Li, Ding, and
  Huang]{yang2021dolg}
Min Yang, Dongliang He, Miao Fan, Baorong Shi, Xuetong Xue, Fu Li, Errui Ding,
  and Jizhou Huang.
\newblock Dolg: Single-stage image retrieval with deep orthogonal fusion of
  local and global features.
\newblock In \emph{IEEE Int. Conf. Comput. Vis.}, pages 11772--11781, 2021.

\bibitem[Zhang et~al.(2021)Zhang, Sattler, and Scaramuzza]{zhang2021reference}
Zichao Zhang, Torsten Sattler, and Davide Scaramuzza.
\newblock Reference pose generation for long-term visual localization via
  learned features and view synthesis.
\newblock \emph{Int. J. Comput. Vis.}, 129\penalty0 (4):\penalty0 821--844,
  2021.

\end{thebibliography}
}

\end{document}